%% file: main.tex
\documentclass[fleqn,10pt]{wlscirep}
\usepackage[utf8]{inputenc}
\usepackage[T1]{fontenc}
\usepackage{float}
\usepackage{appendix}
\usepackage{xcolor}
\usepackage{enumitem,amssymb}
\usepackage{subcaption}
\newlist{checklist}{itemize}{2}
\setlist[checklist]{label=$\square$}
\usepackage{longtable}
\usepackage{soul}
\usepackage{longtable}

\usepackage{url}

\usepackage{breakurl}

\title{In-flight positional and energy use data set of a DJI Matrice 100 quadcopter for small package delivery} 
\author[1,*]{Thiago A. Rodrigues}
\author[2]{Jay Patrikar}
\author[1]{Arnav Choudhry}
\author[3]{Jacob Feldgoise}
\author[4]{Vaibhav Arcot}
\author[1]{Aradhana Gahlaut}
\author[5]{Sophia Lau}
\author[2]{Brady Moon}
\author[6]{Bastian Wagner}
\author[1]{H. Scott Matthews}
\author[2]{Sebastian Scherer}
\author[1,*]{Constantine Samaras}
\affil[1]{Department of Civil and Environmental Engineering. Carnegie Mellon University. 5000 Forbes Avenue, Pittsburgh, 15213, PA USA}
\affil[2]{Robotics Institute, Carnegie Mellon University, 5000 Forbes Avenue, Pittsburgh, 15213, USA}
\affil[3]{Dietrich College of Humanities and Social Sciences, Carnegie Mellon University, 5000 Forbes Avenue, Pittsburgh, PA 15213, USA}
\affil[4]{General Robotics, Automation, Sensing, and Perception Laboratory, University of Pennsylvania, Philadelphia, PA 19104, USA}
\affil[5]{Department of Electrical and Computer Engineering, Carnegie Mellon University, 5000 Forbes Avenue, Pittsburgh, PA 15213, USA}
\affil[6]{Baden-Wuerttemberg Cooperative State University (DHBW), Ravensburg, Germany}

\affil[*]{corresponding author(s): Thiago A. Rodrigues (tarodrig@andrew.cmu.edu), Constantine Samaras (csamaras@cmu.edu)}

\begin{abstract}
We autonomously direct a small quadcopter package delivery Uncrewed Aerial Vehicle (UAV) or "drone" to take off, fly a specified route, and land for a total of 209 flights while varying a set of operational parameters. The vehicle was equipped with onboard sensors, including GPS, IMU, voltage and current sensors, and an ultrasonic anemometer, to collect high-resolution data on the inertial states, wind speed, and power consumption. Operational parameters, such as commanded ground speed, payload, and cruise altitude, are varied for each flight. This large data set has a total flight time of 10 hours and 45 minutes and was collected from April to October of 2019 covering a total distance of approximately 65 kilometers. The data collected were validated by comparing flights with similar operational parameters. We believe these data will be of great interest to the research and industrial communities, who can use the data to improve UAV designs, safety, and energy efficiency, as well as advance the physical understanding of in-flight operations for package delivery drones. 

\end{abstract}
\begin{document}
\flushbottom
\maketitle

\thispagestyle{empty}

\section*{Background \& Summary}
The last decade has seen considerable development towards using Uncrewed Aerial Vehicles (UAVs) or "drones" \cite{stolaroff2018energy} to improve the efficiency, speed, and access of last-mile package delivery. Apart from the ability to carry packages, the majority of these UAVs also have the capability to takeoff and land vertically to reduce their operational footprint and increase their delivery precision. A common solution is to use a multirotor system that generates lift by pushing the air down by using a set of brushless direct current (BLDC) motors that spin counter-rotating propellers. UAVs that produce lift purely using rotors either during the entirety of the mission or for a part of it, need considerably large amounts of energy to lift a package when compared to a conventional winged aircraft where lift is produced by air flowing over a fixed wing. This larger energy requirement has a considerable impact on the range and endurance of these multi-rotor UAVs. The energy required is a function of the environmental factors like the prevailing wind conditions, temperature, humidity, etc; UAV design parameters such as the powerplant specifications, and the shape and build of the UAV; and mission-specific parameters such as the nominal ground speed, operation altitude, and delivery package specifications. Developing energy models that can accurately predict the energy use of these UAVs using these factors as inputs is critical in improving UAV safety and efficiency \cite{zhang2020energy, liu2017power}. Current research uses theoretical models and physical parameters to estimate energy consumption\cite{liu2017power,bezzo2016,troudi2018,abdilla2015,FIGLIOZZI2017251}, while other methods use data-driven approaches to regressed parameters and estimate models\cite{abeywickrama2018empirical,maekawa2017power,prasetia2019mission}.

However, a recent survey \cite{zhang2020energy} on the state-of-the-art of energy modelling for multirotor UAVs compared various models and found that the even with the common sets of parameters, the energy consumption rate (J/m) varies by a factor of 3 to 5 across the models. The survey advocates for comparing results to empirical data from comprehensive drone delivery field tests to improve the energy prediction accuracy. Most studies only conduct a few flight and the raw flight data is not released in public domain to aid comparisons. Given the strict regulatory requirements and significant effort required to conduct UAV field tests, no standard comprehensive data set exists in the public domain to the best of authors' knowledge.     
\par In this work, we performed experiments in order to empirically measure the energy use of a Multirotor UAV with autonomous capabilities while carrying a range of payloads through various campaigns. We autonomously direct a DJI\textsuperscript{\textregistered} Matrice 100 (M100) \cite{DJI} drone to take off, carry a range of payload weights through a triangular flight pattern, and land. We collected high-resolution data on the inertial speeds, altitude, wind and energy for each flight. Between flights, we varied specified parameters, such as commanded ground speed, payload weight, and cruise altitude. We simultaneously collected data from the broad array of on-board sensors throughout 209 flights.

We developed an experimental protocol to ensure the reliability of the data collected. The flights followed a pre-established route with varying altitude (25 m, 50 m, 75 m and 100 m), speed (4 m/s, 6 m/s, 8 m/s, 10 m/s and 12 m/s) and payload mass (no payload, 250 g and 500 g). Each combination was repeated at least three times, totaling 195 flights. In addition, 14 recordings were performed with the drone in hover and idle modes, for a total 209 flights. Finally, the data provided by each sensor were synchronized at a frequency of approximately 5 Hz. Information on the operational setup, such as payload mass, altitude and speed during cruise, date and time the flight started, and the predefined route, was manually logged and attached to the data set.

The flights were performed in the township of Penn Hills, PA at a site  that is approximately 16 kilometers away from Pittsburgh, PA, USA. Proper certifications from federal and local authorities were obtained prior to testing. Checklists were created to ensure that flights would occur as safely and efficiently as expected. Finally, the information collected on each flight was plotted and assessed in order to validate the consistency of the data acquired.

\section*{Methods}
This section details the acquisition setup used in these experiments and the sensor suite mounted on the airframe. It also discusses the experimental protocol for the flight tests. 

\subsection*{Acquisition Setup}
\begin{figure} [H]
    \centering
    \includegraphics[width=\textwidth]{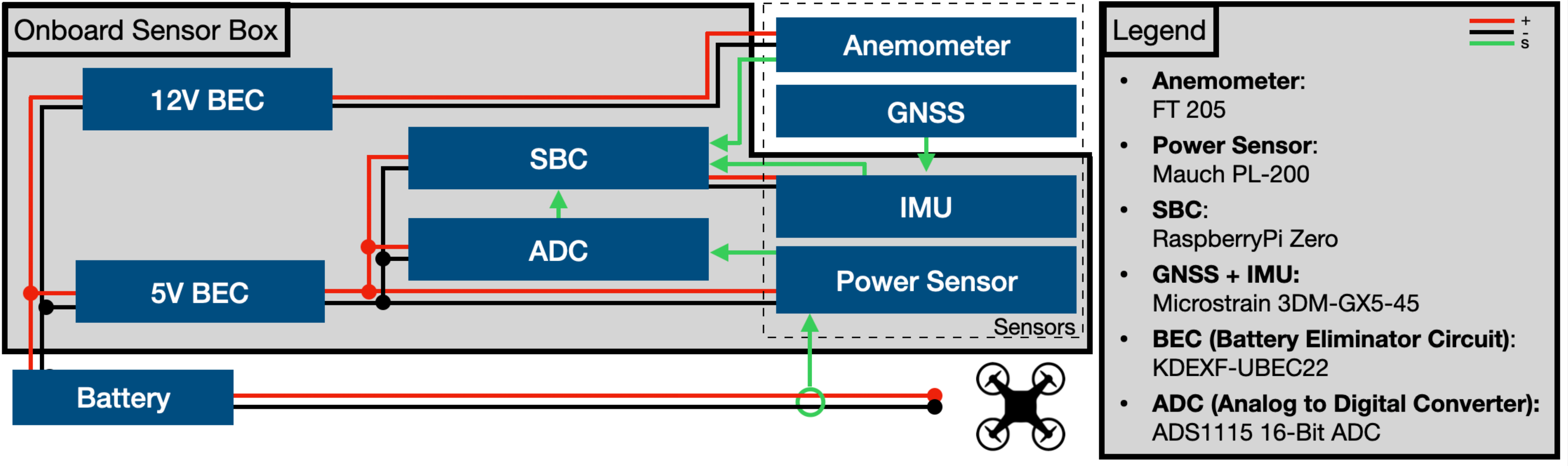}
    \caption{Onboard sensor suite with the complete setup}
    \label{fig:schematic}
\end{figure}

The acquisition setup discusses the hardware and software setup used to collect the data. A complete schematic is shown in Figure \ref{fig:schematic}.

\subsubsection*{Airframe}
\begin{figure} [H]
    \centering
    \includegraphics[width=\textwidth]{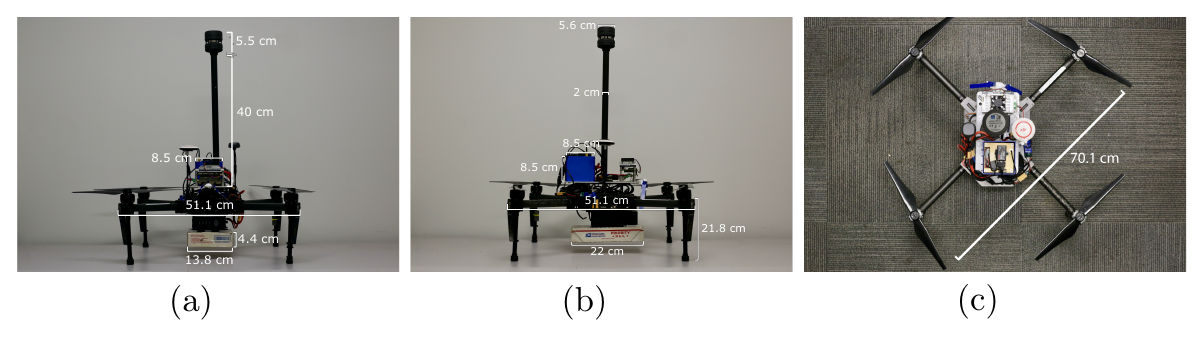}
    \caption{a) Front view, b) Side view, and c) Top view of the DJI Matrice 100 with our payload, sensors and onboard computer. Additional details can be found on the DJI 3D CAD file \cite{djiCAD}}
    \label{fig:dji_pics} 
\end{figure}

Multirotors are characterized by their use of multiple rotors to produce lift. The majority of the commercially available multirotor systems use 3 to 8 rotors to achieve flight. We use the DJI Matrice 100 quadrotor platform to represent multirotor UAVs which is seen in Figure \ref{fig:dji_pics}. The Matrice 100 is a fully programmable and customizable UAS with a maximum cruise speed of 17 m/s (in GPS mode). The airframe is equipped with the DJI 3510 motors (350 Kv), DJI E SERIES 620D ESCs, and we use the DJI 1345s for our rotor blades. The system has an on-board autopilot that provides autonomous capabilities. Its standard battery has a capacity of 4500 mAh which gives it a flight time of 22 minutes without any additional payload. 

The mass of the airframe is 1831 g, the battery 600 g, the anemometer and pole 136g, and the onboard computer with small sensors and wiring 1113 g, totaling 3680 g. Therefore, the total takeoff masses were 3680 g (no payload), 3930 g (with the 250-g payload) and 4180 g (with the 500-g payload). The payloads of different masses were all of dimensions 22 x 13.8 x 4.4 cm and were attached to the bottom of the airframe with velcro straps. The mass of the drone includes the mass of the Velcro straps. 

\subsubsection*{Wind measurement sensor}
\begin{figure}[H]
    \centering
    \includegraphics[width=\textwidth]{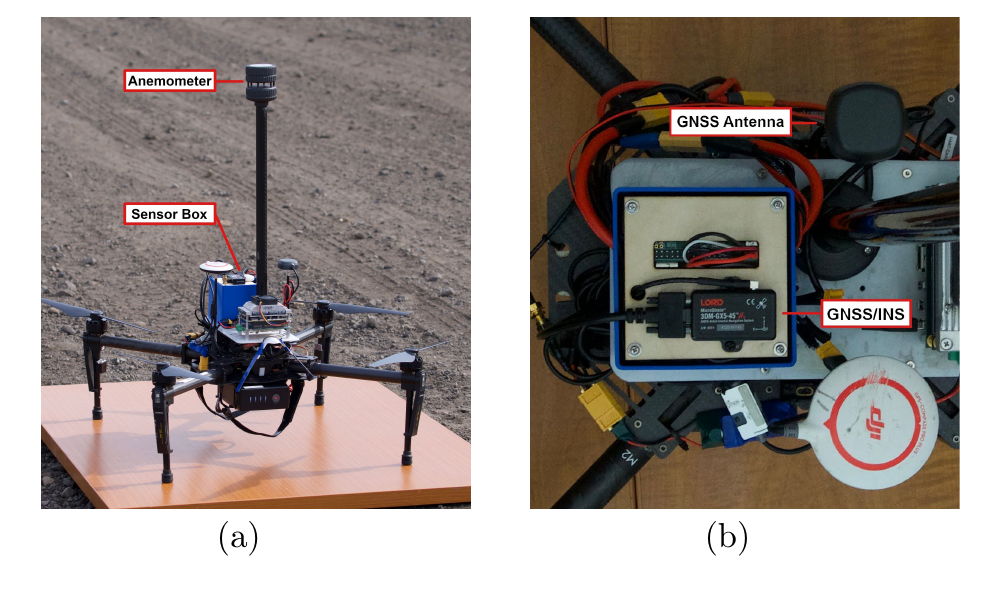}
    \caption{Data acquisition hardware setup. a) DJI Matrice 100 with the anemometer and sensor box attached, b) Top-down view of the platform and sensor box with the 3DM-GX5-45 GNSS INS visible, sensor pack}
    \label{fig:airframe}
\end{figure}

The experiments use a FT Technologies $FT205$ UAV-mountable ultrasonic anemometer for wind measurements \cite{FTTechnologies} which is seen in Figure \ref{fig:airframe}a. The sensor is accurate up to $\pm 0.1 m/s$, has a refresh rate of 10 Hz, and is factory calibrated. We use the device's built-in filtering process to obtain reliable data. UART communication is used to record data from the sensor. 

\subsubsection*{State measurement sensor}
We record the state of the system (position, velocity, and orientation) using the 3DM-GX5-45 GNSS/INS \cite{Lord2018} which is seen in Figure \ref{fig:airframe}b. These sensors use a built-in Kalman filtering system to fuse the GPS and IMU data. The sensor was used at an output rate of 10 Hz. The sensor records data in N(North)-E(East)-D(Down) frame fixed at the takeoff point. The sensor is calibrated as per manufacturer instructions. 

\subsubsection*{Current and Voltage measurement sensor}
The current and voltage supplied to the drone are measured using a Mauch Electronics PL-200 sensor \cite{MAUCH}. This sensor is based on the Allegra ACS758-200U hall current sensor, which can record currents up to 200 A and voltages up to 33 V. The sensor board is only installed into the "positive" (red) main wire from the battery; the "negative" (black) wire stays untouched, which reduces the risk the sensor board might short circuit. A Hall sensor was chosen for its better accuracy when compared with a traditional shunt sensor. The sensor is calibrated as per manufacturer instructions. Analogue readings from the sensor are converted into a digital format using a 8 channel 17 bit analogue-to-digital converter (ADC). The ADC is based on the MCP3424 from Microchip Technologies Inc and is a delta-sigma A/D converter with low noise differential inputs. 

\subsubsection*{Syncing and Recording}

Data syncing and recording is handled using the Robot Operating System (ROS) running on a low-power Raspberry Pi Zero W. Data is recorded on the Raspberry Pi's microSD card. The data provided by each sensor were synchronized to a frequency of approximately 5 Hz using the ApproximateTime \cite{Foundation2010} message filter policy of ROS. The synchronized output approximately follows the current frequency of the sensors. The frequency variability in the synchronized output is thus a by-product of this decision not to interpolate but keep the data intact as received directly from the sensor. All the data has associated timestamps which enables a user to interpolate it at whichever frequency the user desires. A hard synchronization at a particular frequency would have required us to apply an interpolation algorithm, the choice of which we felt should rest with the end-user.

\subsection*{Acquisition Protocol}
This section discusses the flight routines for data collection. A flight plan was created to ensure safety and reliability of the data collected. Procedures for pre-, during, and post-flight were followed as described below. 
\subsubsection*{Pre-flight}

Each test day is selected subject to weather conditions. Extremely harsh weather is avoided, such as precipitation and high wind speeds, but care is taken that the data-set is not biased towards good weather. The test location $(40.465690N,~ -79.788281W)$ is an open field, outside of the urban area. 

The control variables are the programmed speed during cruise (4 m/s, 6 m/s, 8 m/s, 10 m/s, and 12 m/s), programmed altitude during cruise (25 m, 50 m, 75 m, and 100 m), and Payload mass (0, 250 g, 500 g). Flights with all possible combinations of the control variables were carried out to obtain a comprehensive data set. 

Before each flight we ensure that the vehicle is airworthy and that the airframe and control system are calibrated to run the upcoming experiment. We ensure that the flight is configured correctly according to the control variables on the day's flight plan. We then go through the pre-flight checklist (see Supplementary Information). 

\subsubsection*{Flight}
\begin{figure} [H]
    \centering
    \includegraphics[width=\textwidth]{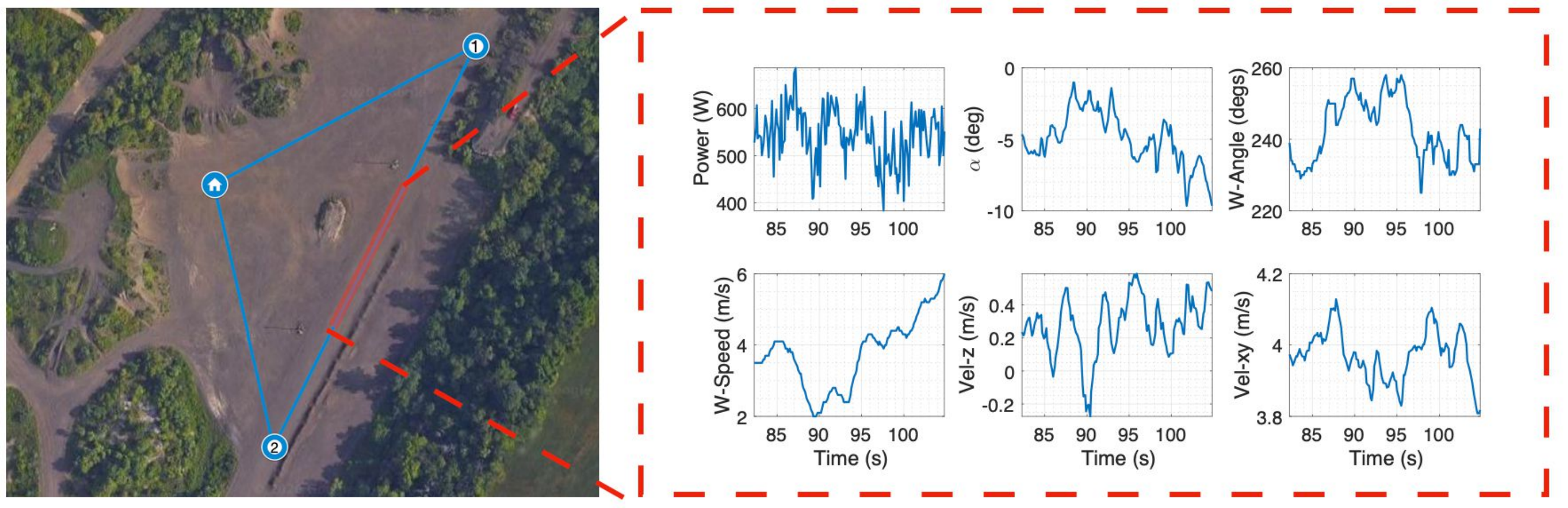}
    \caption{GPS route and sample data outputs from the onboard sensor array.}
    \label{fig:snippet}
\end{figure}

Once the preflight is complete, the Remote Pilot-In-Command (PIC) issues a pre-flight notice to the members and uses a custom Graphical User Interface (GUI) to launch the health monitoring scripts. The program first completes a check of all the onboard sensors and ensures that all subroutines are fully operational. Once the system gets a go-ahead from the health monitoring scripts it then starts the data recording. After a verbal "Go for takeoff" notice, the PIC launches the flight control scripts that use the DJI SDK to autonomously control the GPS waypoint guidance. The UAV arms itself and then takes off vertically to reach the commanded cruise altitude at a constant upward speed of approximately 2.5 m/s. Once at the altitude, the UAV turns the heading to the first waypoint and accelerates to reach the commanded ground speed. On reaching the first waypoint, the UAV stops and turns to go to the next waypoint. This continues as the UAV tracks a triangular wind-neutral path while recording the data. The triangular path and a snippet of the recorded data is shown in Figure \ref{fig:snippet}. When it reaches the launch point again, the UAV stops and descends at approximately 1.5 m/s. Once safely on the ground, the UAV disarms itself and the data recording stops. The PIC calls out flight completed. At all points in the flight, the PIC maintains visual line of sight with the UAV with access to the manual control override to ensure safety of the team and the UAV.  

\subsubsection*{Post-flight}
At the conclusion of the flight, a GUI is used to download the raw ROSBag files and then use the post-processing scripts to synchronize the data from various sensors and convert it to a comma-separated values (CSV) file for storage. The raw files are also stored. Concurrently entries are also completed on the flight plan. The post-flight checklist is followed (see Supplementary Information) and the UAV is then configured for the next flight.

\subsubsection*{Regulation and Safety}
All data collection is conducted within the rules and regulations under part 107 for UAS operations. This includes maintaining visual line-of-sight with the aircraft, keeping the aircraft weight under 55 pounds (24.95 kg), flying only in daylight, not flying above 400 feat above ground level, following a preflight inspection, and flying withing permissible airspace \cite{part107}. It is also ensured that the take-off weight and vehicle commands stay within the airframe safety specification. All FAA and local UAV regulations are observed.

Care is taken in the usage of LiPo DJI TB48D batteries. They are charged using chargers which monitor the voltage and temperature. Batteries are transported in a metal container to protect them from puncture and to contain a fire in the case of combustion. 

Additionally, safety glasses are worn when appropriate and a fire extinguisher was always made available for use.

\section*{Data Records}
All raw and processed data records listed in this section are available at (\href{https://doi.org/10.1184/R1/12683453}{https://doi.org/10.1184/R1/12683453}) \cite{Rodrigues}.The data recorded by each sensor are compiled in a .zip file that contains a folder for each flight with a file named $\textit{raw.bag}$. A CSV file named $\textit{parameters.csv}$ provides a list with all flights and flight parameters. Finally, a file $\textit{flights.zip}$ contains a csv file for each flight with the information on duration, wind speed and direction, current and voltage of the system, aircraft position, orientation speed and acceleration. Table \ref{tab:variables} provides a description on each variable.
\begin{table}[H]
\centering
\caption{\label{tab:variables} Variable description for the data record}
\begin{tabular}{|l|c|l|}
\hline
Variable & Unit/Format & Description \\
\hline
flight &  & An integer that represents the code of the flight performed. \\
\hline
speed & $m/s$  & Programmed horizontal ground speed during cruise. \\
\hline
payload & $g$ & Mass of the payload attached to aircraft. \\
\hline
altitude & $m$ & Programmed altitude. \\
\hline
date & \textit{YYYY-MM-DD} & When the flight was conducted.
  \\
\hline
local\_time & \textit{24:00 h} &  Time of the day when the flight started. \\
\hline
route &  & Predefined path followed by the aircraft.\\
\hline
time & $s$ & time elapsed in flight. \\
\hline
wind\_speed & $m/s$ & Airspeed provided by the anemometer. \\
\hline
wind\_angle & $deg$ & Direction of the air with respect to the north (CW).
 \\
\hline
battery\_voltage & $V$ & System voltage measured immediately after the battery. \\
\hline
battery\_current & $A$ & System current measured immediately after the battery. \\
\hline
position\_x & $deg$ & Longitude of the aircraft.
 \\
\hline
position\_y & $deg$ & Latitude of the aircraft. \\
\hline
position\_z & $deg$ & Altitude of the aircraft with respect to the sea-level. \\
\hline
orientation\_x; \_y;\_z; \_w & $quaternion$ & Aircraft orientation. \\
\hline
velocity\_x; \_y; \_z & $m/s$ & Ground speed. \\
\hline
angular\_x; \_y;\_z & $rad/s$ & Angular rate. \\
\hline
linear\_acceleration\_x; \_y; \_z & $m/s^2$ & Ground acceleration. \\
\hline
\end{tabular}

\end{table}

\section*{Technical Validation}
\begin{figure}[H]
    \centering
    \includegraphics[width=\textwidth]{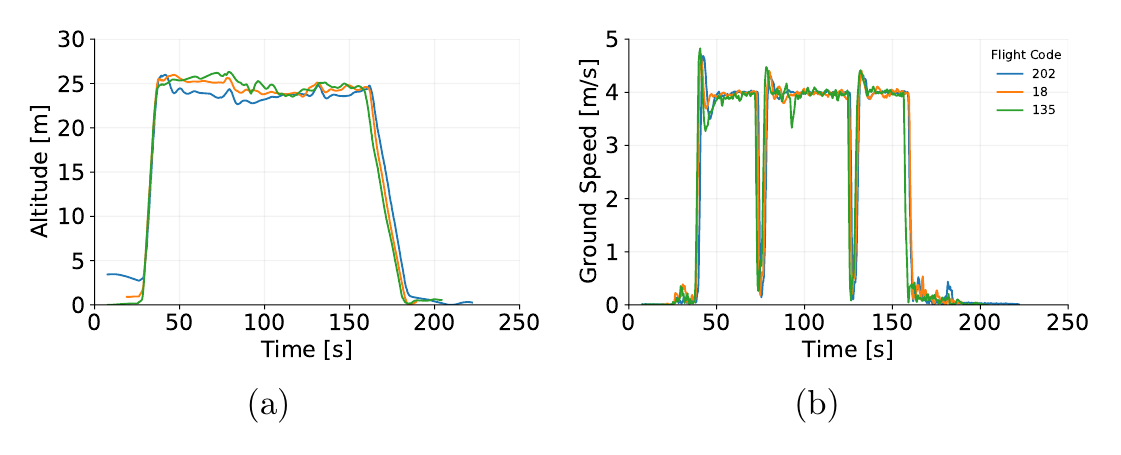}
    \caption{a) Altitude and b) Ground Speed from 3 individual flights operating at cruise speed of 4 m/s, altitude of 25 m and payload of 250 g}
    \label{fig:altitude} 
    
\end{figure}

\begin{figure}[H]
    \centering
    \includegraphics[width=\textwidth]{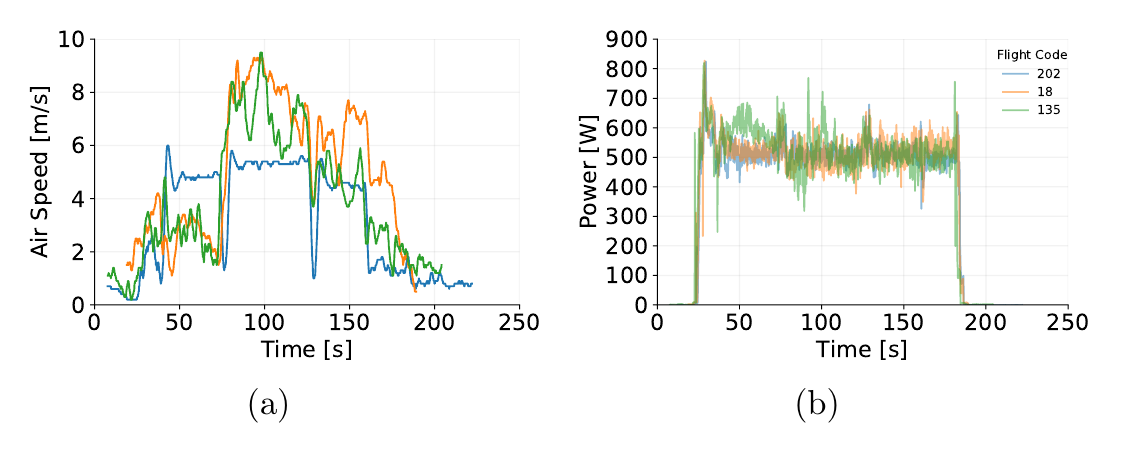}
    \caption{a) Air Speed and b) Power from 3 individual flights operating at cruise speed of 4 m/s, altitude of 25 m and payload of 250 g}
    \label{fig:airspeed} 
    
\end{figure}

\begin{figure}[H]
    \centering
    \includegraphics[width=\textwidth]{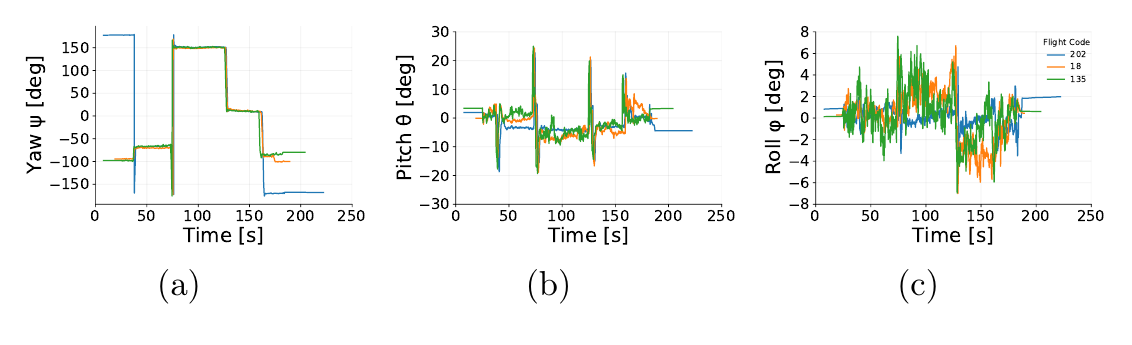}
    \caption{Euler angles from 3 individual flights operating at cruise speed of 4 m/s, altitude of 25 m and payload of 250 g: a) Yaw, b) Pitch and c) Roll}
    \label{fig:euler} 

\end{figure}

The data collected were assessed to ensure the reliability of the data provided. The flights were grouped according to altitude, speed during cruise, and payload. Then, we assessed the parameters collected (positional parameters, wind speed, and power) by comparing flight that share the same setup (programmed altitude, programmed speed and payload mas). Figures \ref{fig:altitude}, \ref{fig:airspeed} and \ref{fig:euler} show examples flights grouped by similar altitude, payload and speed. For instance, flights 18, 135, and 202 show altitude (Figure \ref{fig:altitude}a) and ground speed (Figure \ref{fig:altitude}b) during cruise that oscillate around the programmed parameters of 25 m and 4 m/s, respectively. Figure \ref{fig:airspeed}a shows the influence of wind that naturally varies the airspeed readings among flights. On the other hand, Figure \ref{fig:airspeed}b shows that the power demand is kept consistent for all three flights. The variations in the yaw (Figure \ref{fig:euler}a) at the beginning of the flight shows that the aircraft was initially facing a different direction before the take-off of flight 202. Once the take-off procedure starts the aircraft automatically rotates to face the preset direction, as followed by the other flights. 

\begin{figure} [H]
    \centering
    \includegraphics[width=\textwidth]{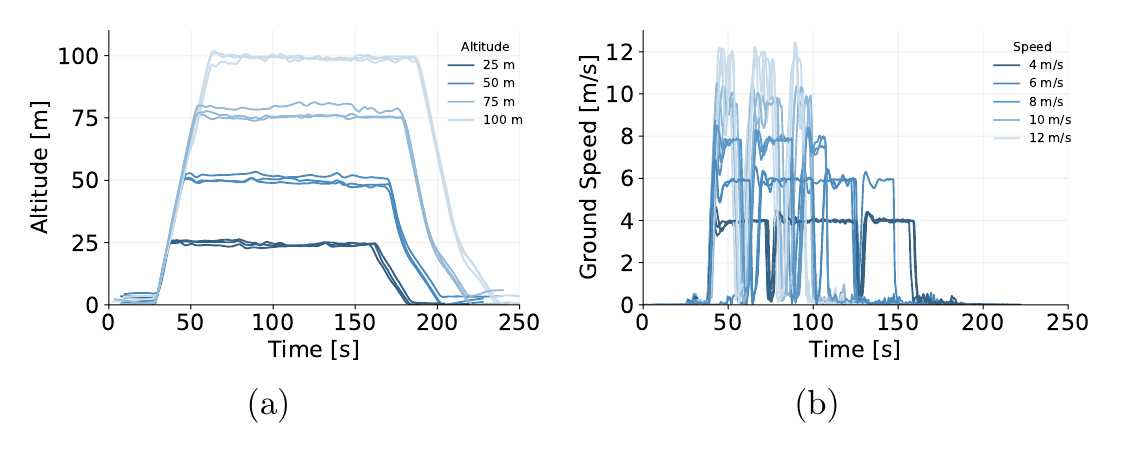}
    \caption{Variation in control variables commanded a) altitude and b) ground speed}
    \label{fig:range} 
    
\end{figure}

\begin{figure} [H]
    \centering
    \includegraphics[width=\textwidth]{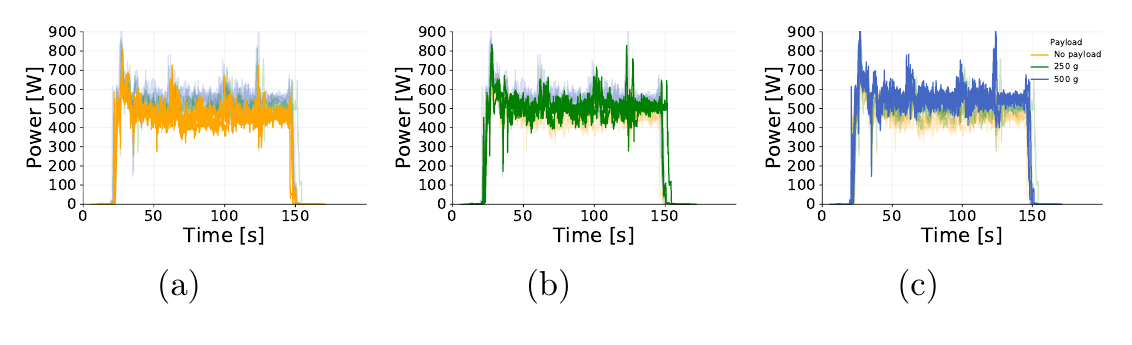}
    \caption{Power readings for 3 individual flights with similar commanded altitude and ground speed but different payloads plotted with emphasis to highlight the difference in power consumption}
    \label{fig:payload} 

\end{figure}

Figures \ref{fig:range} a and b show the range of altitudes and speeds reflected on the data collected, whereas Figure \ref{fig:payload} shows the impact of the payload on the power demand.

The behavior of the parameters assessed shows only minor variations among flights with similar setups. These minor variations were expected due to external factors and the inherent variability of the measuring processes. Nevertheless, in all assessments, the results showed a consistent pattern, with data varying within limits that respect the physical boundaries of the experiment. 

\begin{figure} [H]
    \centering
    \includegraphics[width=\textwidth]{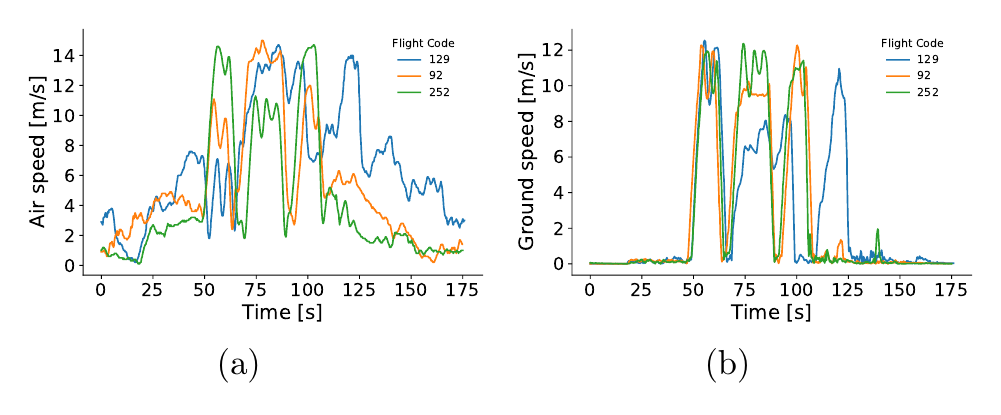}
    \caption{Flight 129 experienced higher wind speed levels and flew at lower speeds when compared to flights with same setup. a) Air speed, and b) ground speed from 3 flights operating at cruise speed of 12 m/s, altitude of 75 m and payload of 500 g}
    \label{fig:group_129} 
\end{figure}

In addition, we computed the total energy consumption of each flight by numerically integrating power (current * voltage) over time. Then, we compared the total energy consumption of flights with the same setup (programmed speed, programmed altitude, and payload mass). The mean relative energy amplitude across the flight groups was 4.3\% with a standard deviation of 2.6\%. Moreover, 95\% of the groups had a relative energy amplitude of less than 10\%, which reflects natural variations among the flights. The maximum energy amplitude was observed by flights 92, 129, and 252 (approximately 3.5 Wh), which represents a relative amplitude of 15.5\% of the mean energy consumption within this group. An in-depth analysis shows that flight 129 (24.8 Wh) had a greater cruise duration, when compared to flights 92 and 252 (21.8 Wh and 21.3 Wh, respectively). However, the air speed experienced by flight 129 after 100 seconds (Figure \ref{fig:group_129}a) might have reduced the actual ground speed of the drone (Figure \ref{fig:group_129}b), increasing the total flight duration, and causing this discrepancy on the overall energy consumption. 

A similar analysis with two other groups of flights that had relative energy amplitude greater than 10\% did not indicate any substantial difference among the flight recorded. 

We have also implemented a script to automatically process and assess the values measured during flight and compare them to the programmed values of altitude. 173 out of the 194 flights with cruise movement did not have any altitude readings during cruise beyond the manufacturer's $\pm$ 5m tolerance range. Moreover, there was only one flight that had a mean altitude during cruise (107.1m) beyond the programmed altitude and manufacturer's range (105.0 m). However, as the flight pattern and energy consumption of this flight was consistent with the similar flights (relative energy amplitude of 3\% within the group) we have decided to maintain its data in the data set.

Finally, the uncertainty inherent to each sensor used has been summarized on Table \ref{tab:tolerances} according to the values provided by each manufacture.  

\begin{longtable}{ll}
\caption{Manufacturer's accuracy for each sensor}
\label{tab:tolerances}\\
\textbf{Parameter}             & \textbf{Accuracy} \\ \hline
\endfirsthead
\endhead
\hline
\endfoot
\endlastfoot
Vertical position {[}m{]}      & $\pm$ 5           \\
Horizontal position {[}m{]}    & $\pm$ 2           \\
Velocity {[}m/s{]}             & $\pm$ 0.1         \\
Roll, Pitch, Heading {[}deg{]} & $\pm$ 2.0         \\
Wind speed {[}m/s{]}           & $\pm$ 0.3         \\
Wind direction {[}deg{]}       & $\pm$ 4           \\
Current {[}\%{]}               & $\pm$ 1.2       \\ \hline
\end{longtable}

\section*{Usage Notes}
The data available can be used to model the energy consumption of a small quadcopter drone, empirically fitting the results found or validating theoretical models. These data can also be used to assess the impacts and correlations among the variables presented and/or the estimation of non-measured parameters, such as drag coefficients. These data should not be extrapolated to assess the energy consumption of different drone models or drones operating outside of the range of values tested.

The measurements from the onboard anemometer can be used to calculate the airspeed components (cross-wind and head-wind) and local wind conditions. The anemometer records the wind angle and magnitude with respect to the moving drone. We provide the original data from the anemometer. However, previous works that have used anemometers for onboard wind measurements \cite{garibaldi2010uav,bruschi2016wind,thorpe2018measurement} have reported that while the wind angle measurements were found to be accurate for all flight conditions, the magnitude measurements show a bias that reports a higher wind magnitude than expected. We found that the effect is more pronounced at lower UAV speeds. Thus, future work is required to investigate and correct the magnitude bias in addition to correcting for the UAV motion. In Supplementary Information, we provide one method of magnitude bias and ego motion correction, although work in this area is ongoing. 

\section*{Code availability}
The codebase used to fly the UAV autonomously, record and synchronize data, and interface between different sensors is available for public use under BSD license. The codebase can be accessed at \href{https://bitbucket.org/castacks/workspace/projects/DOE}{https://bitbucket.org/castacks/workspace/projects/DOE}. 

\bibliography{main}

\section*{Acknowledgements} 
This work was supported by the U.S. Department of Energy’s Vehicle Technologies Office, Award Number DE-EE0008463. This material is also based upon work supported by the National Science Foundation Graduate Research Fellowship under Grant No. DGE1745016. The work was also supported by LORD MicroStrain Sensing Systems.

\section*{Author contributions statement}
The research project was conceived by C.S, H.M. and S.S. The experiment was designed by J.P. The experiments were conducted by  T.R., J.P., A.C., J.F., V.A., A.G and B.M. The results were analysed by T.R. All authors reviewed the manuscript. 

\section*{Competing interests}
The authors declare no competing interests.

\newpage
\appendix
\input{appendix}

\end{document}

%% file: appendix.tex


\section{Checklists}
\subsection{Pre-Flight}
\label{apen1:pre}

\begin{checklist}
    \item Inspect the aircraft 
    \begin{checklist}
        \item Visually inspect the condition of the unmanned aircraft system components
        \item Inspect the airframe structure, including undercarriage, all flight control surfaces and linkages
        \item Inspect registration markings for proper display and legibility
        \item Inspect moveable control surface(s), including airframe attachment point(s)
        \item Inspect servo motor(s), including attachment point(s)
        \item Inspect the propulsion system, including powerplant(s), propeller(s), rotor(s), ducted fan(s), etc.
    \end{checklist}
    \item Verify all systems (e.g. aircraft, control unit) have an adequate energy supply for the intended operation and are functioning properly
    \item Inspect the avionics, including control link transceiver, communication/navigation equipment and antenna(s)
    \item Check UAS compass and calibrate UAS compass prior to any flight if necessary.
    \item Check that the display panel, if used, is functioning properly
    \item Check ground support equipment, including takeoff and landing systems, for proper operation
    \item Check on board navigation and communication data links
    \item Check flight termination system and manual override switch
    \item Check battery levels for the aircraft and control station
    \item Check that the payload is securely attached
    \item Verify communication with UAS and that the UAS has acquired GPS location from at least 4 satellites
    \item Arm and disarm the UAS propellers to inspect for any imbalance or irregular operation
    \item If required by flight path walk through, verify any noted obstructions that may interfere with the UAS
    \item At a controlled low altitude, fly manually within range of any interference and recheck all controls and stability
\end{checklist}

\subsection{Post-Flight}
\label{apen1:post}

\begin{checklist}
    \item Wait for all motors and rotors to stop
    \item Verify stop sensor recording
    \item Disconnect the battery
    \item Inspect the airframe for any in-flight damage
    \item Check wiring is secure
    \item Check connectors are fully connected
    \item Motors are securely mounted
    \item Propellers are securely fastened
    \item Propellers rotate freely without any obstructions (e.g. hitting edges of incorrectly fitted indoor hull) or have been removed.
    \item All sensors are unobstructed and undamaged
\end{checklist}

\section{Anemometer Magnitude Bias and Ego Motion Correction}
\label{chap:bias}

\begin{figure} [H]
    \centering
    \includegraphics[width=0.45\textwidth]{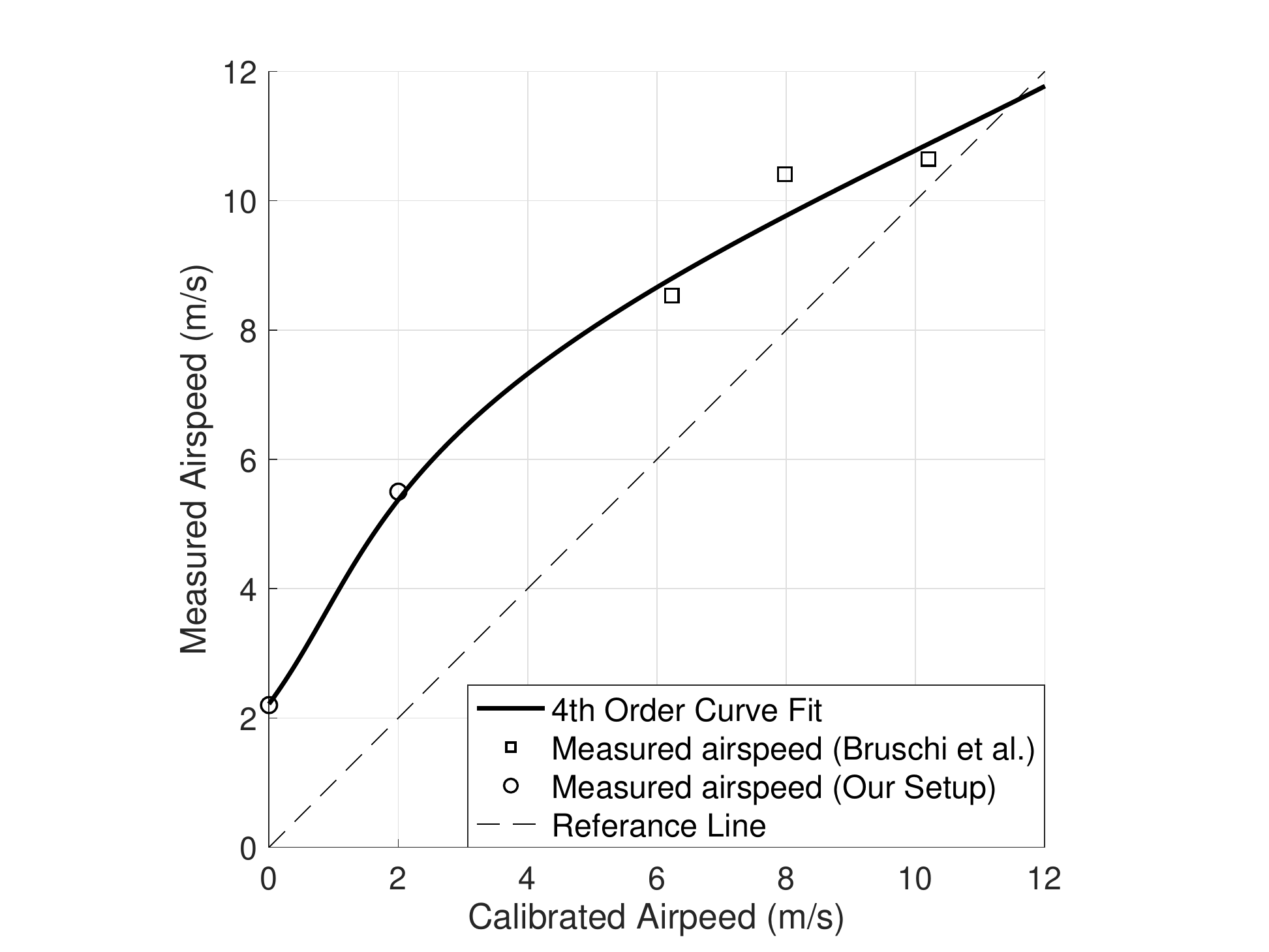}
    \caption{Anemometer corrections}
    \label{fig:anemo_usage}
\end{figure}

One way to remove the magnitude bias and ego motion in the wind measurements is by using the inertial velocity data and the data collected by Brushi et al. \cite{bruschi2016wind} of a quadrotor flying with a ultrasonic wind sensor inside a wind tunnel. Given the similarity between our setup and theirs, the wind-tunnel results\cite{bruschi2016wind} are used to construct a polynomial mapping between measured and actual wind speeds. As the data lacks points in the lower range of values, it is augmented with results from our own setup. The data-points and the curve fit are plotted in Figure \ref{fig:anemo_usage}. The UAV was flown in hover and constant ground speed mode for an extended amount of time in near-zero wind conditions to get the lower range of values. The magnitude correction polynomial and ego-motion correction is given in Equation \ref{eq:corr}.
\begin{equation}
    \begin{split}
w_m^{[actual]} &= -0.002(w_m^{[raw]})^4 + 0.052(w_m^{[raw]})^3-0.421(w_m^{[raw]})^2 + 1.917w_m^{[raw]} - 2.7\\
w_N &= (w_m^{[actual]})\cos(w_{\theta}^{[raw]}) - V_N\\
w_E &= (w_m^{[actual]})\sin(w_{\theta}^{[raw]}) - V_E\\
w_m^{[corrected]} &= \sqrt{w_N^2 + w_E^2}\\
w_{\theta}^{[corrected]} &= \arctan(w_E,w_N)\\
\end{split}
\label{eq:corr}
\end{equation}
where $V_N$ and $V_E$ represent the inertial speed of the UAV in North and East directions, and $w_N$ and $w_E$ represent the corrected wind components in North and East directions. 

The data set contains the raw measurements and is not modified with these proposed methods, as the focus of this work is not about obtaining accurate wind field estimations. The area of wind estimation through the use of a drone is an ongoing field of research\cite{Patrikar-2020-125335}, as the rotors induce complex air disturbances which affect the anemometer. Thus we left our data in a state where various current or future methods of wind correction can be applied if needed.
